\PassOptionsToPackage{unicode}{hyperref}
\PassOptionsToPackage{hyphens}{url}
\documentclass[
  11pt,
]{article}
\usepackage{xcolor}
\usepackage[margin=1in]{geometry}
\usepackage{amsmath,amssymb}
\usepackage{iftex}
\ifPDFTeX
  \usepackage[T1]{fontenc}
  \usepackage[utf8]{inputenc}
  \usepackage{textcomp} % provide euro and other symbols
\else % if luatex or xetex
  \usepackage{unicode-math} % this also loads fontspec
  \defaultfontfeatures{Scale=MatchLowercase}
  \defaultfontfeatures[\rmfamily]{Ligatures=TeX,Scale=1}
\fi
\usepackage{lmodern}
\ifPDFTeX\else
\fi
\IfFileExists{upquote.sty}{\usepackage{upquote}}{}
\IfFileExists{microtype.sty}{% use microtype if available
  \usepackage[]{microtype}
  \UseMicrotypeSet[protrusion]{basicmath} % disable protrusion for tt fonts
}{}
\makeatletter
\@ifundefined{KOMAClassName}{% if non-KOMA class
  \IfFileExists{parskip.sty}{%
    \usepackage{parskip}
  }{% else
    \setlength{\parindent}{0pt}
    \setlength{\parskip}{6pt plus 2pt minus 1pt}}
}{% if KOMA class
  \KOMAoptions{parskip=half}}
\makeatother
\usepackage{longtable,booktabs,array}
\usepackage{calc} % for calculating minipage widths
\usepackage{etoolbox}
\makeatletter
\patchcmd\longtable{\par}{\if@noskipsec\mbox{}\fi\par}{}{}
\makeatother
\IfFileExists{footnotehyper.sty}{\usepackage{footnotehyper}}{\usepackage{footnote}}
\makesavenoteenv{longtable}
\usepackage{graphicx}
\makeatletter
\newsavebox\pandoc@box
\newcommand*\pandocbounded[1]{% scales image to fit in text height/width
  \sbox\pandoc@box{#1}%
  \Gscale@div\@tempa{\textheight}{\dimexpr\ht\pandoc@box+\dp\pandoc@box\relax}%
  \Gscale@div\@tempb{\linewidth}{\wd\pandoc@box}%
  \ifdim\@tempb\p@<\@tempa\p@\let\@tempa\@tempb\fi% select the smaller of both
  \ifdim\@tempa\p@<\p@\scalebox{\@tempa}{\usebox\pandoc@box}%
  \else\usebox{\pandoc@box}%
  \fi%
}
\def\fps@figure{htbp}
\makeatother
\providecommand{\tightlist}{%
  \setlength{\itemsep}{0pt}\setlength{\parskip}{0pt}}
\usepackage{newunicodechar}
\newunicodechar{⁶}{\ensuremath{^6}}
\newunicodechar{⁻}{\ensuremath{^-}}
\newunicodechar{→}{\ensuremath{\rightarrow}}
\newunicodechar{−}{\ensuremath{-}}
\newunicodechar{∝}{\ensuremath{\propto}}
\newunicodechar{≥}{\ensuremath{\geq}}

\usepackage{float}
\let\origfigure\figure
\let\endorigfigure\endfigure
\renewenvironment{figure}[1][2]{\origfigure[H]}{\endorigfigure}
\usepackage{bookmark}
\IfFileExists{xurl.sty}{\usepackage{xurl}}{} % add URL line breaks if available
\hypersetup{
  hidelinks,
  pdfcreator={LaTeX via pandoc}}

\author{}
\date{}

\begin{document}

\section{Confident at the moment of action: belief miscalibration in LLM
play under hidden
information}\label{confident-at-the-moment-of-action-belief-miscalibration-in-llm-play-under-hidden-information}

Bhushan Kashinath Joshi

bjoshi2779@yahoo.com

\subsection{Abstract}\label{abstract}

Agentic systems increasingly gate actions on a model's own stated
confidence, which assumes confidence tracks correctness at the moment of
acting. We test this in a hidden-information chess variant where royal
status can be secretly, repeatedly relocated between pieces, and where
an agent's stated probability distribution over the opponent's hidden
royal piece --- elicited every turn, separately from the move it chooses
--- is scored against ground truth recoverable after the game. Across
two independent batches, captures made at high stated confidence (≥0.5)
about the hidden piece's location were correct in 1 of 62 cases. The
calibration deficit is concentrated almost entirely in these events:
99.3\% of it in the original batch, 98.7\% in the replication. The same
pattern, in weaker form, orders consistently (point estimates only; most
pairwise gaps are not statistically distinguishable at this sample size)
across four further model configurations spanning a second provider ---
reported as scope for the finding, not as evidence that capability
predicts calibration: a same-model comparison at a fixed external
leaderboard score shows a deliberation-budget change alone moves the
metric by nearly as much as a large cross-model gap. In a separate seat,
conventional evaluation axes --- legality, cost, latency, completion
rate --- can dissociate entirely from belief quality, with the
configuration winning on every conventional axis producing the worst
belief quality tested. A model exhibiting this pattern can still win the
game its belief was about, which is why outcome-only evaluation would
not detect it.

\begin{center}\rule{0.5\linewidth}{0.5pt}\end{center}

\subsection{1. Introduction}\label{introduction}

Agentic deployments increasingly gate consequential actions on a model's
own stated confidence --- escalate to a human only if the model reports
itself below some threshold, act autonomously only above it. This
pattern assumes that a model's stated confidence, at the moment it acts,
tracks whether that action is actually correct. That assumption is not
safe in general: a recent large-scale benchmark of verbalized confidence
across frontier models finds that its most accurate model is not its
best-calibrated one, confirming accuracy and calibration measure
genuinely different things even before any hidden-state task is involved
{[}10{]}. Whether the assumption holds under hidden information
specifically is rarely tested directly, because doing so requires two
things most evaluation settings don't provide together: a moment where
the model states a belief about something not directly observable, and a
distinct action taken at the same operational point, elicited
independently of the stated belief. This permits belief quality to be
evaluated at the action-selected locus, and belief--action
correspondence to be analysed separately --- both checked against a fact
that can be recovered independently of either.

Standard evaluations cannot test this assumption because they collapse
the two moments into one. A QA benchmark scores an answer; a verbalized-
confidence study scores that same answer's stated probability of being
correct. There is no separate action to check the stated belief against
--- the belief \emph{is} the graded output. A competitive-game
benchmark, by contrast, has an action (a move) but typically no elicited
belief about hidden state at all, and scores only the game's outcome.
Neither setting can show a model stating one thing and doing another,
because neither elicits a belief and an action as two separable events.

We build a setting where both are observable and ground truth for the
hidden state is exact and recoverable at reveal: a chess variant in
which royal status can be secretly, repeatedly relocated between pieces,
and in which we elicit, every turn, both a move and a probability
distribution over where the opponent's hidden royal piece currently is.
Because the true answer is logged internally and never exposed to either
player during play, we can score the stated belief against fact at any
point, and compare it against the move chosen that same turn.

Doing so, we find that a model's stated belief and its subsequent action
come apart precisely at the moment the action is taken: when the model
captures a piece at high stated confidence about where the hidden royal
piece is, it is correct in only 1 of 62 such cases across two
independent batches --- a gap concentrated almost entirely in these
high-confidence events, not spread evenly across the model's beliefs in
general. Critically, a model exhibiting exactly this pattern can still
win the game the belief was about, which is precisely why an evaluation
that scores only outcomes does not surface it.

Our contributions: (1) a measurement of stated-belief miscalibration at
the moment of action, replicated across two independent batches with a
consistent mechanism, and scoped --- not claimed universal --- across
five seat-configurations spanning two model providers; (2) a
demonstration, using a separate seat, that conventional evaluation axes
(legality, cost, latency, completion rate) can dissociate entirely from
the capability this setting is built to measure, with the configuration
that wins on every conventional axis producing the worst belief quality
tested; and (3) a setting design --- detailed in §3 and discussed for
its own methodological contribution in §6 --- in which a hidden fact is
created and revised by the agent under test rather than assigned to it
externally, and in which a source-visible opponent makes it possible to
bound what any predictor could achieve before concluding a model failed
to achieve it.

\begin{center}\rule{0.5\linewidth}{0.5pt}\end{center}

\subsection{2. Related Work}\label{related-work}

\textbf{Game-based LLM evaluation.} Contamination and saturation have
pushed LLM evaluation toward competitive games, where correctness cannot
be memorized from a static answer key. Kaggle Game Arena launched with
chess in August 2025 {[}1{]} and had, by early 2026, expanded to include
poker and Werewolf alongside chess and Four in a Row {[}2, 3{]}. Each
game carries its own leaderboard and metric --- chess uses an Elo rating
fit by Bradley--Terry over all-play-all pairwise outcomes within that
game {[}2{]} --- with a unified ranking aggregating per-game scores into
a single consolidated figure {[}2{]}. In every case the ranking is over
game outcomes, not over the quality of the reasoning that produced them.
Game Arena's launch post frames the effort against the difficulty of
knowing ``if models trained on internet data are actually solving
problems or just remembering answers they've already seen,'' and against
benchmark saturation, where near-ceiling scores stop revealing
meaningful performance differences {[}1{]}. Its Chess Openings variant
seeds each game from one of the twenty most popular two-ply openings in
Lichess data, addressing an observed tendency of models in the original
text-chess format to default to a narrow set of memorized lines (the
Sicilian Defense chief among them); openings play is used to separate
genuine calculation from recalled theory {[}4{]}. We adopt the same
contamination-resistance rationale for choosing a synthetic,
non-standard ruleset --- but every metric across Game Arena's games is
an \emph{outcome} metric: win, loss, chip stack, vote share. None of it
elicits a model's stated belief about hidden state and scores that
belief against ground truth, which is the layer our setting is built to
measure.

\textbf{LLM calibration and uncertainty.} A separate literature asks
whether a model's verbalized confidence tracks the correctness of its
own answers on QA-style tasks, from the finding that models can predict
``P(IK)'' --- the probability they know an answer --- reasonably well in
the right format, though calibration of that prediction degrades on
unfamiliar tasks {[}5{]}, to recent work showing verbalized confidence
is highly sensitive to elicitation protocol: which answer string is
scored and under what conditioning context, with the sign of the ECE
comparison left undetermined by protocol choice alone in half of one
paper's twelve tested configurations {[}6{]}. Other recent work targets
overconfidence directly: ConfTuner fine-tunes models against a tokenized
Brier-score loss, proven a proper scoring rule, to produce
better-calibrated verbal confidence {[}7{]}, and DINCO estimates and
corrects for a model's suggestibility bias by having it verbalize
confidence across self-generated distractor claims, then combines the
resulting normalized confidence with a self-consistency signal for a
further calibration gain {[}8{]}. A mechanistic strand traces the
failure further upstream: RLHF is shown to push models toward verbalized
overconfidence specifically because the reward models used in its
optimization step carry an inherent bias toward high-confidence scores
independent of response quality {[}9{]}, and a recent 15-model benchmark
scoring verbalized confidence with the same Brier-score methodology used
throughout our own analysis finds five of fifteen models, spanning three
labs, score worse than a calibrated-random baseline {[}10{]} ---
evidence the pattern is not one lab's idiosyncrasy. The same benchmark
also names Gemini 3.1 Flash-Lite --- this paper's own primary seat (§3)
--- the single worst-calibrated model it tested, by a wide margin
{[}10{]}; we return to what that means for this paper's own finding in
§5. Closest to our own setting, recent work on long-horizon agents in
partially observable environments separates an agent into a belief-state
model and a policy model, representing the belief as atomic claims
carrying verbalized certainty labels and conditioning the policy on that
compact belief rather than the full interaction history {[}11{]}. That
separation is an architectural choice made to improve task performance
and bound context growth, and it presumes the verbalized certainty
attached to each claim is informative enough to act on. Our setting
inverts the question: rather than designing a decoupling, we measure one
that was not intended --- whether a model's own stated belief and its
chosen action agree at the moment the action is taken --- and find that
they need not. In the QA literature the stated belief and the scored
answer are the same channel, so belief--action consistency is
structurally unmeasurable there; in Agent-BRACE the belief state is
optimized against downstream task reward, never scored against the
action it informed, so it remains unmeasured. A related strand asks not
whether a single stated confidence is calibrated but whether belief
updates coherently as multi-turn evidence accumulates, comparing
per-turn model distributions against exact Bayesian posteriors where one
exists and asking whether better latent inference translates into better
downstream prediction {[}12{]}; their measurement points are
researcher-chosen turns and their downstream object is a prediction,
whereas Regent Chess's capture-time analysis conditions calibration on
realized capture events within an ongoing game. Bearing most directly on
this question, concurrent work finds that LLM actions frequently
contradict separately elicited confidences across prediction-market
betting, tool invocation, and user-challenge settings, naming the
pattern the action--belief gap {[}13{]} --- we do not claim novelty for
the general observation that stated belief and chosen action can
diverge. What differs: they ask whether a subsequent action is
\emph{coherent} with an elicited confidence, while we ask whether a
stated probability is \emph{empirically calibrated} --- whether cases
assigned probability \emph{p} are correct at approximately frequency
\emph{p} --- against independently logged ground truth at the
action-selected locus, and a model can be action-consistent yet badly
miscalibrated, or well-calibrated yet act incoherently; they elicit
confidence in a static setting and observe behaviour in an interactive
one, their own framing, whereas Regent Chess elicits the hidden-state
distribution within the same structured response as the move, inside the
evolving game, rather than importing a confidence measured on a separate
static task; and their targets are future events or static facts, while
ours is a hidden state present at every point, exactly recoverable
afterward, and relocatable by the opponent mid-game. Our own
free-control result (§4.7) --- isolated-position elicitation overstating
in-game performance by 28.4 points on one seat and 41.2 on another ---
is itself a caveat on static elicitation generally, including theirs,
not a criticism of their result.

\textbf{Theory-of-mind and hidden-information benchmarks.} LLM agents
have been placed in imperfect-information games explicitly to probe
opponent modeling and deception: WOLF, a Werewolf-based evaluation
presented at the NeurIPS 2025 Workshop on Multi-Turn Interactions in
Large Language Models, separately measures deception production and
detection across role-grounded agents, finding werewolves produce
deceptive statements on 31\% of turns while peer detection reaches only
71--73\% precision and roughly 52\% overall accuracy --- production
comes considerably more easily than detection, scored against the
speakers' own self-reported honesty as ground truth {[}14{]}, a weaker
standard than the logged, model-independent fact our own setting scores
against (§3). A related instrument, MafiaScope, probes agent beliefs
after every public utterance in the social-deduction game Mafia, on a
context copy whose answers never re-enter gameplay, scores first-order
role beliefs against ground truth the engine holds, and relates those
beliefs to subsequent votes {[}15{]} --- the closest prior instrument in
kind, though, as in Werewolf, the state being predicted is a role
assigned to another agent at setup, not one the tracked agent itself
creates and revises. A separate study of extended-play Texas Hold'em
finds theory-of-mind-like opponent modeling emerges over repeated hands,
but only when agents are given persistent memory {[}16{]}. In every such
setting, the hidden state is \emph{exogenous} --- a role assigned at
setup, a hand of cards dealt once --- fixed and outside the agent's
control for the game's duration; the inference task is entirely about
someone \emph{else's} fixed secret. Our setting inverts this: the hidden
fact (which piece secretly holds royal status) is chosen and can be
relocated by the player \emph{themself}, mid-game, unobserved, making
concealment a matter of managing a self-created, revisable intention
rather than protecting a static fact one did not choose.

\begin{center}\rule{0.5\linewidth}{0.5pt}\end{center}

\subsection{3. Setting and Method}\label{setting-and-method}

\textbf{Regent Chess.}\footnote{The game variant used as this paper's
  instrument was designed by the author prior to this work; AI
  assistance (Claude, chat interface) was used for its software
  implementation and to surface rule edge cases during that
  implementation. Rule correctness is established by a test suite
  covering each rule clause, written with AI assistance and verified by
  the suite's execution.} We use a digital chess variant in which royal
status can be secretly transferred between pieces. The objective is not
fixed to the king: a player wins by capturing the opponent's
\emph{Regent}, the piece that currently holds royal status, which may or
may not be the Original King. A \emph{Crown Shift} --- the secret action
of transferring royal status to another piece --- becomes available to
each player at the start of their 4th turn and again 15 moves after each
use; it is executed silently, with no declaration to the opponent, and
does not consume the player's turn (a regular move must still follow).
Outside of Crown Shift, standard chess rules apply. Full rules are given
in the Appendix, reproduced from the frozen ruleset under which every
reported game was played.

The design property this paper depends on: \textbf{ground truth is exact
and recoverable at reveal.} Every Crown Shift is logged internally
(invisible to both players and to the agent under test during play) and
can be replayed after the game to recover, for any ply, which piece was
the true Regent. This is what permits scoring a stated belief against a
fact, not a proxy.

\textbf{Elicitation.} Each real turn, the agent under test receives the
board state, its own move history, and its own known state (own color,
own current Regent, own mode) --- never the opponent's --- and returns a
single structured response: a chosen move, and a probability
distribution over which of the opponent's live pieces is most likely to
be the opponent's Regent, expressed as a top-\emph{k} list of (square,
probability) pairs with an explicit residual mass for ``some other
opponent piece.'' The prompt explicitly names the mover's own color and
the opponent's color and states that named squares must hold an opponent
piece, never the mover's own --- a disambiguation fix found to matter
substantially for some seats (§4.7). We flag here, not only in
Limitations, that this is a top-\emph{k} elicitation: a model can fail
to name a square for reasons ranging from ``did not consider it'' to
``considered it, but it did not make the cut'' to ``the format asked for
fewer candidates than the model's actual uncertainty spanned'' --- the
coverage finding in §4.6 must be read with this caveat live, not
appended afterward.

\textbf{Opponent.} All games are played against a fixed, deterministic
heuristic agent (Tier-1, medium difficulty) that does not model the
LLM's beliefs and follows a fixed, source-visible policy. A constant,
non-adaptive opponent is what makes seat-to-seat comparison and
depth-indexed curves meaningful (a stronger or better-calibrated
opponent would confound ``the model got worse'' with ``the opponent got
better''), and a \emph{source-visible} policy is the precondition for
the oracle-vs-public tractability bound discussed in §6 --- neither is
available against an opponent whose policy cannot be read from code.

\textbf{Scoring.} We score belief quality two ways, treated as separate
questions throughout, never blended: \textbf{capture-only} scoring pairs
the model's stated probability on a square with whether that square held
the true Regent, at the moment the model captures a piece on that square
--- an exposure-biased population by construction (a piece is captured
partly because it is capturable, which anti-correlates with actually
being the hidden Regent). \textbf{Every-ply} scoring pairs every named
candidate square, on every well-formed ply, with whether it is the true
Regent that ply --- richer, an order of magnitude more events, and free
of capture-time exposure bias. Both are scored against a
\textbf{population-matched uniform prior} (1 / number of the opponent's
live pieces, computed per-ply, not a single reused constant) as the
baseline, using Brier score with paired bootstrap confidence intervals
(same event indices resampled for both curves under comparison). The
pre-shift window (board-ply \textless{} 8) is excluded from every
belief-quality comparison: the rules guarantee the Original King is
still the Regent until each side's 4th turn, so during that window the
correct answer is knowable by rule alone, and any predictor that has
read the rules gets a free, structurally unfair advantage over one that
has not (§4.2).

\textbf{Pre-registration.} Four expectations were written before any
model played, each with a stated falsifier, and scored verbatim rather
than reworded to fit results: that top models would not excel initially
and would instead produce fluent but poorly-calibrated belief
statements; that illegal-move rates would rise with game length; that
stated beliefs would show a belief--action gap; and that tracking
accuracy and narrative quality would show a possible asymmetry, which
required an elicitation format never built and is recorded as not
scoreable rather than force-fit. The first three are directly the
results reported in §4.1--§4.2, §4.5, and §4.6 respectively. All
headline figures in this paper were re-derived from version-controlled
analysis code immediately before writing, after an earlier internal
check found a figure that had not been re-verified against a
subsequently fixed bug in an unrelated tracking component.

\textbf{Seats.} We refer to each model-plus-sampling-configuration
pairing under test as a \emph{seat} (e.g., a specific model at a
specific temperature and token budget). Seats are labeled in the order
they were audited for this work, not in the order they appear in any
table or figure --- several seats audited early were later dropped from
calibration scoring for reasons specific to them (excess truncation, an
adapter incompatibility, or a well-formed rate too low to yield
scoreable events) and appear only where a result of theirs is directly
relevant (§4.4, by model name; §4.7, by label), never in Table 1; this
is why the labels below are non-consecutive. Five seat-configurations
across two model families/labs were scored for capture-time calibration
(Table 1). Only one seat (S1, Gemini 3.1 Flash-Lite) received the full
calibration battery (capture-only, every-ply, and ground-truth-scored
belief quality across many games) --- every other seat's number is
capture-time high-confidence hit rate only, from a smaller sample. This
asymmetry is stated here and repeated at the point each cross-seat
figure is used (§4.3), not left implicit.

\textbf{Table 1.} Seat configurations, external leaderboard scores,
sampling parameters, and calibration-scoring sample sizes for the five
seats compared in §4.3.

\begin{longtable}[]{@{}
  >{\raggedright\arraybackslash}p{(\linewidth - 10\tabcolsep) * \real{0.1667}}
  >{\raggedright\arraybackslash}p{(\linewidth - 10\tabcolsep) * \real{0.1667}}
  >{\raggedright\arraybackslash}p{(\linewidth - 10\tabcolsep) * \real{0.1667}}
  >{\raggedright\arraybackslash}p{(\linewidth - 10\tabcolsep) * \real{0.1667}}
  >{\raggedright\arraybackslash}p{(\linewidth - 10\tabcolsep) * \real{0.1667}}
  >{\raggedright\arraybackslash}p{(\linewidth - 10\tabcolsep) * \real{0.1667}}@{}}
\toprule\noalign{}
\begin{minipage}[b]{\linewidth}\raggedright
Seat
\end{minipage} & \begin{minipage}[b]{\linewidth}\raggedright
Model
\end{minipage} & \begin{minipage}[b]{\linewidth}\raggedright
Kaggle chess-text leaderboard score
\end{minipage} & \begin{minipage}[b]{\linewidth}\raggedright
Temperature
\end{minipage} & \begin{minipage}[b]{\linewidth}\raggedright
\texttt{max\_tokens}
\end{minipage} & \begin{minipage}[b]{\linewidth}\raggedright
Games (calibration-scored)
\end{minipage} \\
\midrule\noalign{}
\endhead
\bottomrule\noalign{}
\endlastfoot
S1 & gemini-3.1-flash-lite & \emph{no external anchor} & 0.7 & 4096 & 80
(62 high-conf. capture events) \\
S7 & gpt-5-mini-2025-08-07 & 622.59 & \textbf{1.0 (API-forced)} & 4096 &
--- (14 high-conf. capture events) \\
S6-base & gemini-3-flash-preview & 1258.32 & 0.7 & 4096 & --- (24
high-conf. capture events) \\
S6B & gemini-3-flash-preview & 1258.32 & 0.7 & 16384 & --- (10
high-conf. capture events) \\
S5 & gemini-3.1-pro-preview & 1367.40 & 0.7 & 4096 & --- (8 high-conf.
capture events) \\
\end{longtable}

The ``score'' column is this leaderboard's own reported metric label,
not an Elo rating (retrieved 2026-08-15); we use its scale directly
rather than relabel it. S1 has no entry on the external leaderboard used
for the other four seats. S7 is the only non-Google seat scored here,
and its sampling temperature is an API-enforced deviation from the rest
of the table (\texttt{gpt-5-mini} rejects any value other than 1.0), not
a discretionary choice --- stated here because it sits directly on the
claim S7 is used to support (§4.3), not filed only as a footnote.

\begin{center}\rule{0.5\linewidth}{0.5pt}\end{center}

\subsection{4. Results}\label{results}

\subsubsection{4.1 Primary finding: capture-time miscalibration is real,
replicated, and concentrated
(S1)}\label{primary-finding-capture-time-miscalibration-is-real-replicated-and-concentrated-s1}

We measured S1's capture-time calibration in two independent batches,
run under a pre-registered replication design rather than pooled from
the start: in the original batch, \textbf{0 of 22} captures made at ≥0.5
stated confidence were correct; in a separate, later-run replication
batch, \textbf{1 of 40} were. The replication did not just add sample
size --- it \emph{agreed in direction and magnitude} with the original
(Brier ratio vs.~matched uniform: 7.127 original vs.~7.094 replication,
a 95\% CI on the replication of {[}4.298, 13.135{]}, about 26\% tighter
than the original's {[}3.825, 15.740{]}). Pooling both batches:
\textbf{1 of 62} high-confidence captures were correct (1.6\%).

At capture time specifically, S1's Brier score is 0.1445 against a
matched uniform prior's 0.0203 --- a paired-bootstrap 95\% CI on the
ratio of \textbf{3.8×--15.9×}. The uniform comparator is
population-matched --- recomputed per ply over the opponent's live
pieces --- but is not conditioned on the exposure mechanism by which a
position becomes a capture opportunity; it is a matched uninformative
baseline, not a model of action-conditioned difficulty. This gap is not
diffuse: \textbf{99.3\% of the total Brier gap between S1 and uniform is
concentrated in the high-confidence events above} (98.7\% in the
replication batch, independently); the remaining, lower-confidence
events contribute essentially nothing to the miscalibration. Figure 1
(the reliability diagram) shows this directly --- S1's capture-only
curve sits at or near zero hit rate across nearly the entire confidence
range, with the exception of a small positive rate at the extreme top
bin.

\subsubsection{4.2 The gap is not confined to capture time --- it holds
from the first ply where inference is
possible}\label{the-gap-is-not-confined-to-capture-time-it-holds-from-the-first-ply-where-inference-is-possible}

Restricting to every well-formed ply from board-ply ≥ 8 onward
(excluding the rule-determined pre-shift window, §3), S1's stated
beliefs are \textbf{significantly worse than a population-matched
uniform prior}, in both batches independently at ply≥8 (delta +0.059
original batch, +0.070 replication batch), growing worse at stricter ply
cutoffs in both (to +0.098--+0.112 at ply≥14). This is a separate
population from §4.1's capture-only result (§3's every-ply scoring, not
capture-only), but it points the same direction: the failure is not
confined to the instant of action, though it is far more severe there.
We note explicitly that an earlier internal analysis pooling all plies
including the pre-shift window found no gap in either direction (``tied
with uniform''); that reading did not survive excluding the
rule-determined window, where a model that has read the rules gets a
free, structurally unfair advantage over a baseline that has not. We
report only the corrected, ply≥8 comparison here.

\subsubsection{4.3 Scope: the failure orders consistently across five
seat-configurations from two
labs}\label{scope-the-failure-orders-consistently-across-five-seat-configurations-from-two-labs}

\begin{longtable}[]{@{}llll@{}}
\toprule\noalign{}
Seat & Capture-time hits / n & Rate & 95\% Wilson CI \\
\midrule\noalign{}
\endhead
\bottomrule\noalign{}
\endlastfoot
S1 & 1 / 62 & 1.6\% & {[}0.3\%, 8.6\%{]} \\
S7 & 2 / 14 & 14.3\% & {[}4.0\%, 39.9\%{]} \\
S6-base & 7 / 24 & 29.2\% & {[}14.9\%, 49.2\%{]} \\
S6B & 4 / 10 & 40.0\% & {[}16.8\%, 68.7\%{]} \\
S5 & 4 / 8 & 50.0\% & {[}21.5\%, 78.5\%{]} \\
\end{longtable}

Point estimates order monotonically and span two labs (Google, OpenAI);
see Figure 2. We state plainly what this ordering does and does not
support: pairwise significance clears only at the two extremes (S1
vs.~S5); the three middle points, and S7 against all three of its
neighbors, are not pairwise distinguishable at these sample sizes.
\textbf{We report the ordering as scope evidence that the failure is not
confined to one model; we do not claim external capability predicts
calibration}, and one internal comparison argues directly against that
stronger claim: S6 and S6B are the \emph{same model at the same external
leaderboard score} (1258.32), differing only in deliberation budget
(\texttt{max\_tokens} 4096 vs.~16384), yet their capture-time rates
differ by 10.8 points (29.2\% vs.~40.0\%) --- comparable in size to the
14.9-point gap between S7 and S6-base, which spans a real, substantial
635.7 points on that leaderboard's scale. A configuration change within
one model moves this metric by about as much as a large cross-model
capability gap does.

Two further caveats on this ordering, stated here rather than deferred:
S7's sampling temperature is forced to 1.0 by its API, unlike every
other scored seat's 0.7 (§3) --- S7's point estimate is not a clean
seat-only comparison. And the two most capable seats scored (S5, S6/S6B)
are the same model family and lab; ``capable models are less prone to
this failure'' rests, at present, on evidence from one lab, in the same
way the failure claim itself rested on one lab before S7 was added. On
the narrower question of whether the underlying overconfidence pattern
itself is specific to one seat --- the objection our own five points can
only partially answer at this n --- independent evidence points the same
direction: a 15-model verbalized-confidence benchmark using the same
Brier-score methodology we use throughout finds general verbalized
overconfidence pervasive across frontier systems, not an artifact of any
single lab or model size {[}10{]}.

\subsubsection{4.4 A second pillar: conventional evaluation metrics
dissociate from the capability this paper
measures}\label{a-second-pillar-conventional-evaluation-metrics-dissociate-from-the-capability-this-paper-measures}

In earlier smoke-pilot work with a separate seat (DeepSeek V4 Flash, not
part of the calibration comparison above), disabling the model's
extended reasoning mode produced a configuration that scored best on
every conventional axis a standard LLM evaluation would report: 100\%
model-driven legal moves, 0\% illegal attempts, 0\% truncated responses,
and roughly 28× cheaper and 25× faster per game than the
reasoning-enabled configuration of the same model. That same
configuration's well-formed belief rate was 3.8\%. The configuration
that would win a conventional benchmark on every reported axis is, on
this evidence, the one demonstrably not tracking the board. This is not
confined to one seat's idiosyncrasy: it is the within-model version of
this paper's central argument --- that legal, fluent, cheap, fast play
and accurate belief-tracking are not the same capability and do not move
together, and an evaluation that only scores the former will select for
the latter's absence.

\subsubsection{4.5 Depth-dependent format degradation, and a residual
specific to hidden-state
tracking}\label{depth-dependent-format-degradation-and-a-residual-specific-to-hidden-state-tracking}

S1's well-formed elicitation rate (the rate at which its response parses
as a valid move plus a valid belief distribution) declines with game
depth: 85.6\% in the opening third of a game (agent-own-ply 1--25) to
47.8\% at mid-depth (26--50) to 41.0\% deep in the game (51+, though
this bucket is under-powered, n=39). Illegal move-attempt rate rises
over the same range, from 1.2\% (opening) to 4.5\% (mid-depth) to 20.0\%
(deep) --- a roughly 17-fold increase.

To ask whether this decline is specific to tracking hidden state, we ran
a control: the identical seat, harness, and elicitation format, against
the identical opponent, with the hidden-Regent mechanic disabled
entirely (standard chess, no Crown Shift). The control declines too ---
from 85.8\% (opening) to 65.7\% (mid-game), a 20.1-point drop --- which
rules out ``hidden-state tracking is the sole cause'' of Regent Chess's
own decline (a 37.8-point drop over the same range, from 85.6\% to
47.8\%): if hidden-state tracking were the only mechanism, a condition
with no hidden state at all should not decline nearly this much. It does
not, however, mean the two conditions decline at the same rate ---
Regent Chess's decline is substantially steeper.

A direct per-bucket significance test confirms this is real, not noise:
at ply 1--25, the two conditions are statistically indistinguishable (p
= 0.907); at ply 26--50 --- the only bucket where both conditions carry
enough samples to compare reliably (n = 228 vs.~671; the deep bucket's n
= 39 vs.~538 is both under-powered and confounded by which games survive
that long) --- Regent Chess sits significantly below the control, 47.8\%
vs. 65.7\% (p = 1.6×10⁻⁶). The honest reading is neither
``hidden-state-specific'' nor ``general context growth alone'': general
context growth accounts for real, substantial decline even with no
hidden state to track (the control's own 20.1-point drop), and a
further, independently measurable, hidden-state-specific residual
accounts for the remaining difference between how far each condition
fell from its own opening rate --- 37.8 points for Regent Chess, 20.1
for the control, a 17.7-point difference-in-declines.\footnote{This
  17.7-point figure is close to but not identical to the raw 17.9-point
  gap between the two conditions' levels at ply 26--50 alone (65.7\% −
  47.8\%); both are correct for what they represent --- 17.7 is a
  difference-in-declines (accounting for each condition's own opening
  rate), 17.9 is a level-gap at one matched depth --- and they differ
  only because the two conditions' opening rates are not themselves
  exactly equal (85.6\% vs.~85.8\%).} We scope this claim explicitly to
the single well-powered bucket where it is measurable, not to depth in
general.

\subsubsection{4.6 Coverage: most captures target a square the model
never named as a
candidate}\label{coverage-most-captures-target-a-square-the-model-never-named-as-a-candidate}

These are two distinct properties and are reported separately
throughout: action-time calibration (§4.1--§4.2) asks whether stated
confidence tracks correctness at the moment of action; belief--action
consistency, measured here, asks whether the square acted upon featured
in the stated belief at all. Neither implies the other.

Restricting to captures where that same ply's belief distribution named
two or more candidates (n = 91), the square S1 actually captured had
been named as \emph{any} candidate only 26.4\% of the time (24/91) ---
statistically indistinguishable from a pre-registered, exposure-matched
permutation null of 29.2\% {[}24.2\%, 35.2\%{]} (p = 0.218, one-sided).
We state the elicitation caveat here, not as an afterthought: this is a
top-\emph{k} format (§3), and a model can fail to name a square for
reasons that are not all ``did not consider it.'' What is real and
convention-free regardless of that caveat: \textbf{73.6\% of S1's
captures target a square it never listed as a candidate at all.} Among
the smaller subset where the eventually-captured square \emph{was} named
(n = 24), its rank among the named candidates was in fact high (mean
percentile 0.788, top-ranked 70.8\% of the time) --- the opposite
direction from an earlier, since-withdrawn characterization of this
result as a systematic rank-inversion. We flag this figure's own limit
rather than let it stand unqualified: 17 of these 24 events named only a
single candidate, making ``top-ranked'' close to tautological for most
of the subset (there is only one square to rank), so the 70.8\% figure
should be read as a weak, directionally-suggestive check against the
withdrawn inversion claim, not as positive evidence of good ranking on
its own. The gap that survives --- and is not subject to this limitation
--- is a candidate-coverage gap, not a ranking failure.

\subsubsection{4.7 Free-form screening overstates real-game
performance}\label{free-form-screening-overstates-real-game-performance}

A static, single-position elicitation test --- the kind a lightweight
pre-deployment check would plausibly use --- measurably overstates how a
seat performs across a real, multi-turn game. For S1, a free,
isolated-position control measured 89.3\% well-formed; the same seat,
same prompt, inside real games measured 60.9\% --- a 28.4-point gap. For
a second seat (S2), the gap was larger still: 53.6\% free-control
vs.~12.4\% in real games (a pre-registered stop-rule halted further
spend on that seat once this gap was confirmed), 41.2 points. The gap is
not fixed, and is worse for the weaker of the two seats tested. A
single-position screening test is not a substitute for measuring inside
the interaction it is meant to predict.

\subsubsection{4.8 Pre-registered expectations,
scored}\label{pre-registered-expectations-scored}

Of the four expectations stated in §3, three could be scored against
these results and one could not. The first (fluent but poorly-calibrated
belief statements, rather than early excellence) is confirmed by
§4.1--§4.2. The second (illegal-move rate rising with game length) is
confirmed by §4.5's 1.2\%→4.5\%→20.0\% progression. The third (a
belief--action gap) is confirmed by §4.6, exactly as stated --- not the
stronger rank-inversion form considered and withdrawn during analysis
(§4.6's own note); the confirmed result takes the specific form of a
coverage gap, a special case of the action--belief divergence documented
more broadly elsewhere {[}13{]}. The fourth, an asymmetry between
tracking accuracy and narrative quality, required an elicitation format
that was never built in this work and is recorded as not scoreable
rather than force-fit to adjacent results.

\begin{center}\rule{0.5\linewidth}{0.5pt}\end{center}

\subsection{Figures}\label{figures}

\begin{figure}
\centering
\pandocbounded{\includegraphics[keepaspectratio,alt={Reliability diagram: stated confidence vs.~hit rate at capture time, pooled across both scored batches (n=394 capture-only events), with the corrected reference tracker curve. Marker area is proportional to bin n; no line is drawn across an empty bin. Cited in §4.1.}]{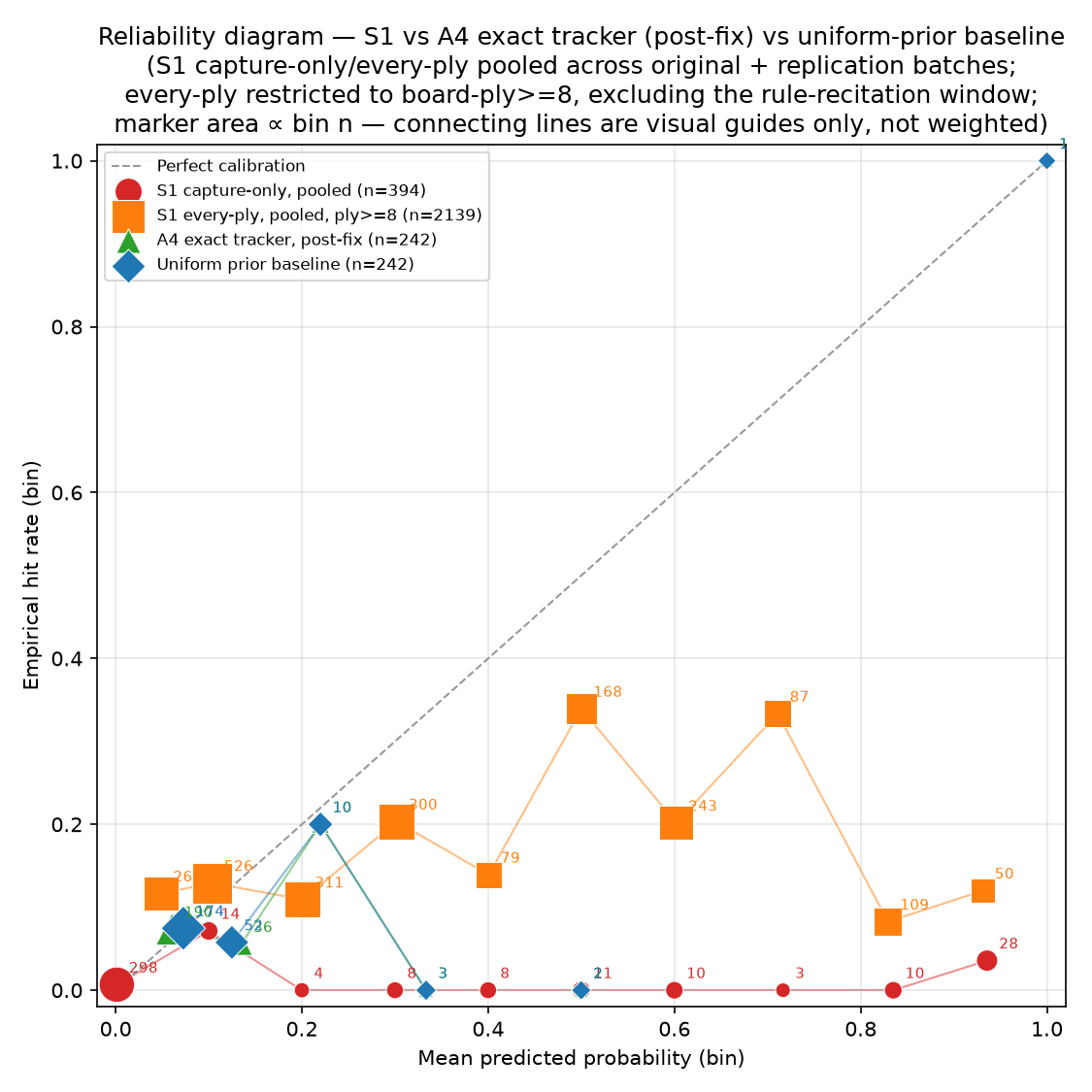}}
\caption{Reliability diagram: stated confidence vs.~hit rate at capture
time, pooled across both scored batches (n=394 capture-only events),
with the corrected reference tracker curve. Marker area is proportional
to bin n; no line is drawn across an empty bin. Cited in §4.1.}
\end{figure}

\begin{figure}
\centering
\pandocbounded{\includegraphics[keepaspectratio,alt={Five-seat capture-time calibration ordering, with 95\% Wilson confidence intervals. Cited in §4.3.}]{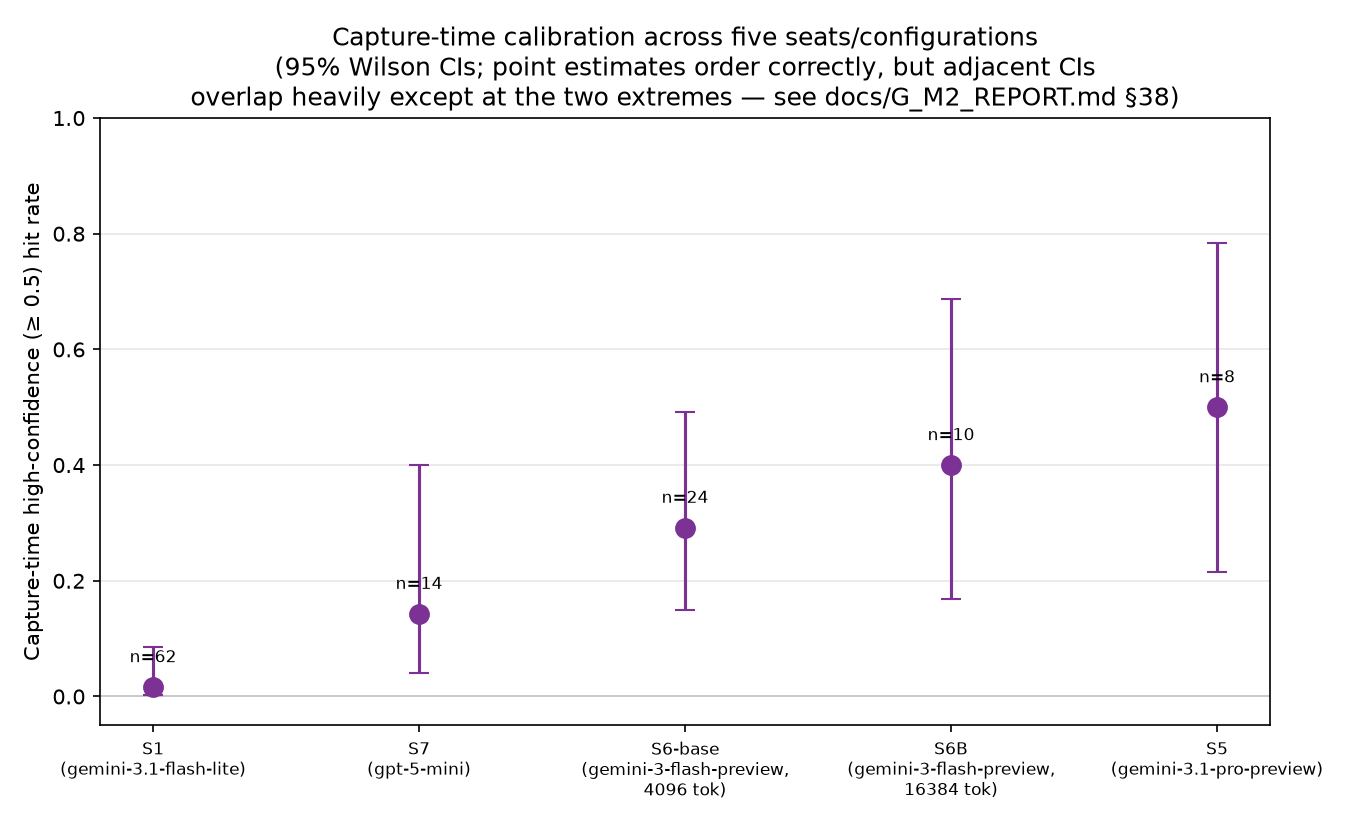}}
\caption{Five-seat capture-time calibration ordering, with 95\% Wilson
confidence intervals. Cited in §4.3.}
\end{figure}

\begin{figure}
\centering
\pandocbounded{\includegraphics[keepaspectratio,alt={Well-formed-rate degradation curves by ply bucket, Regent Chess vs.~the no-crown control. Cited in §4.5.}]{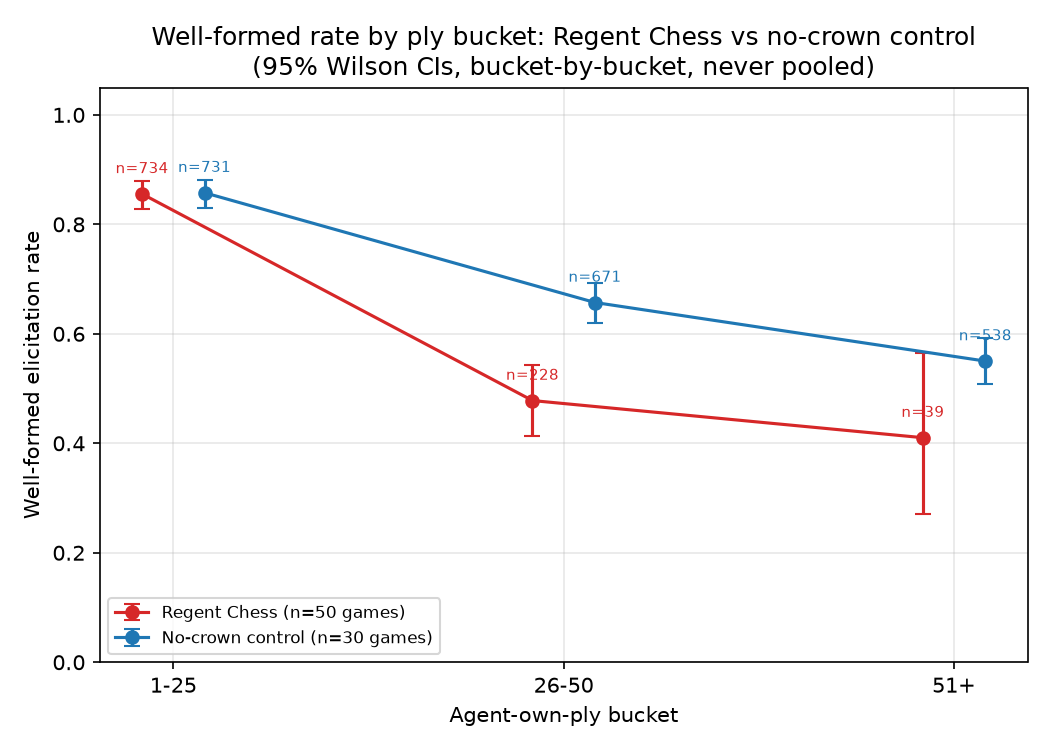}}
\caption{Well-formed-rate degradation curves by ply bucket, Regent Chess
vs.~the no-crown control. Cited in §4.5.}
\end{figure}

\begin{figure}
\centering
\pandocbounded{\includegraphics[keepaspectratio,alt={Supplementary: leaderboard score vs.~capture-time calibration. Cited narratively in §4.3 (the S6/S6B same-score point); does not stand alone as evidence for any claim.}]{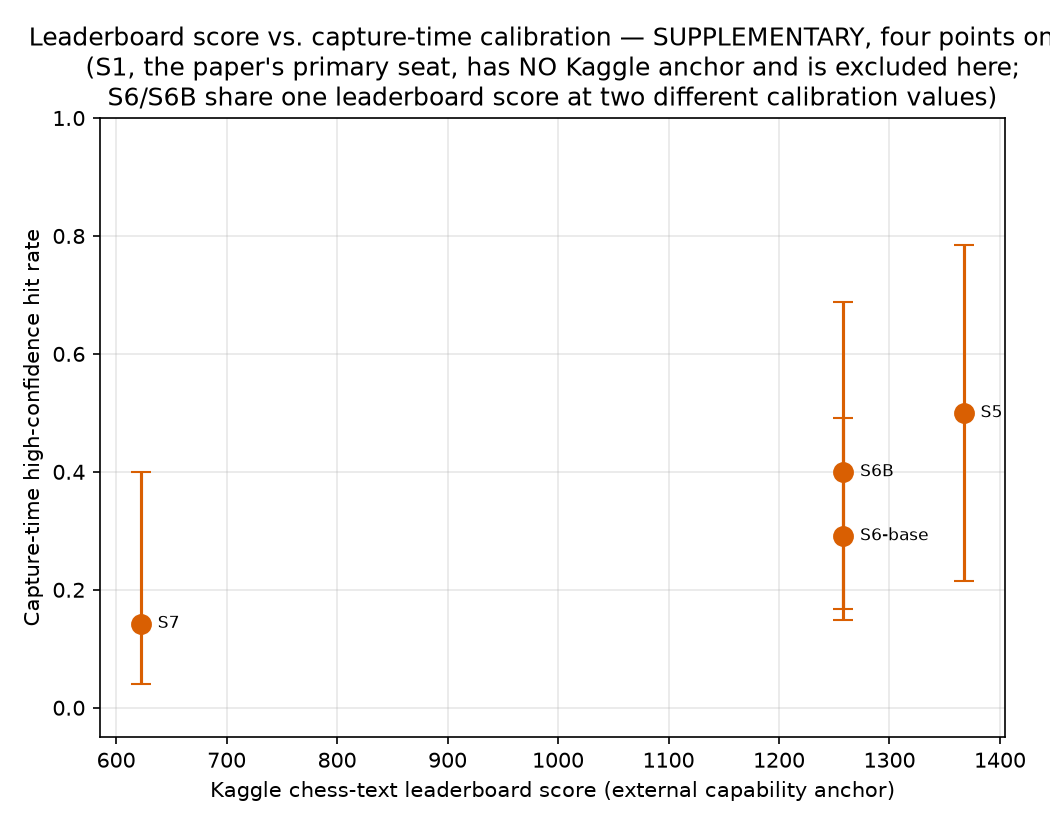}}
\caption{Supplementary: leaderboard score vs.~capture-time calibration.
Cited narratively in §4.3 (the S6/S6B same-score point); does not stand
alone as evidence for any claim.}
\end{figure}

\begin{center}\rule{0.5\linewidth}{0.5pt}\end{center}

\subsection{5. Limitations}\label{limitations}

\textbf{Single seat, full battery.} Only S1 received the complete
calibration battery (capture-only, every-ply, ground-truth-scored belief
quality across many games). The every-ply worse-than-uniform result
(§4.2), the capture-time Brier ratio and its mechanism (§4.1), the
coverage figure (§4.6), and the depth-degradation curves (§4.5) are all
S1-only findings. The five-seat ordering (§4.3) supports one narrow
slice of scope --- capture- time high-confidence hit rate --- and
nothing about every-ply calibration, coverage, or depth effects in the
other four seats.

\textbf{A chess-capability leaderboard score is not this paper's only
external anchor, and a within-model confound undercuts the ordering's
strongest reading.} S1 does not appear on the Kaggle chess-text
leaderboard used to place the other four seats (§4.3), and that
leaderboard's score is a proxy for chess capability, not calibration ---
largely orthogonal to what this paper measures. A more relevant anchor
exists: ConfidenceBench {[}10{]} independently scores Gemini 3.1
Flash-Lite's general verbalized-confidence calibration and finds a
bimodal confidence distribution (mass near a 25\% floor or in an
80--100\% band) that resembles the pattern we observe at capture time
(§4.1, Figure 1) --- corroboration from a different instrument, not
something we constructed. It also covers S5 (gemini-3.1-pro-preview).
Separately, S6 and S6B --- the same model at the same external
leaderboard score --- differ in capture-time calibration by 10.8 points
from a deliberation-budget change alone, comparable to a cross-model gap
spanning 635.7 points on that leaderboard's scale. Together these mean
the ordering in §4.3 cannot support a claim that external capability
predicts calibration on this task, only that the failure generalizes
across labs.

\textbf{Is this just a generally badly-calibrated model, not a
hidden-state- tracking failure?} ConfidenceBench {[}10{]} independently
names Gemini 3.1 Flash-Lite the worst-calibrated model of fifteen
tested, by a wide margin --- a real objection, since it raises the
possibility that our finding merely restates a known, general property
of this model rather than anything specific to belief-tracking under
hidden information. General miscalibration predicts uniformly poor
calibration across contexts; it does not predict what we actually
observe: 99.3\% (98.7\% in replication) of the Brier gap concentrated in
a narrow band of high-confidence capture events rather than spread
evenly across S1's stated beliefs (§4.1), nor does it predict the
pattern attenuating monotonically across five model/ configuration seats
spanning two labs, including seats ConfidenceBench never evaluated at
all (§4.3). A generally overconfident model would look bad everywhere;
what we find instead is a model that looks fine most of the time and is
confidently, specifically wrong at the moment it acts on a hidden-state
belief. We also note S1 was selected for this paper's pilot work on cost
grounds, before and independent of any knowledge of its calibration
profile on any benchmark --- this finding was not obtained by searching
for a badly-calibrated model to demonstrate the effect.

\textbf{Single opponent, single game length regime.} All results are
against one fixed, moderate-difficulty heuristic opponent. The
depth-degradation residual (§4.5) is measurable only in one ply bucket
against this opponent; whether it holds at greater depth, or against a
stronger or position-responsive opponent, is untested.

\textbf{No transfer validation.} Nothing in this work establishes that
performance on this instrument predicts performance on any real-world or
downstream task. Every number here should be read as domain-internal
until that validation exists.

\textbf{Sampling and reproducibility.} S1's provider rejects a fixed
random seed outright, so its games are not exactly replayable from a
recorded seed --- only re-runnable under the same configuration; the
replication reported in §4.1--§4.2 used exactly this re-run design, and
its agreement with the original batch is itself evidence the finding is
not an artifact of one particular random draw. Provider-default,
non-deterministic sampling applies throughout; sampling parameters are
reported in Table 1 where they matter to a specific claim.

\textbf{Footnote-level caveats} (stated briefly; do not individually
threaten a headline claim): truncation rates vary substantially and
uncontrolled across seats (from near-zero to over 80\% at a fixed token
budget), and a seat that truncates more often is not on equal footing
with one that does not; colour balance across real-game batches was
checked and found balanced (40--60\% white/black) for every seat scored
here; sampling non-determinism means exact byte-for-byte reproduction of
a specific game is not always possible, only reproduction of the
aggregate finding under the same configuration; the top-\emph{k}
elicitation caveat on the coverage finding (§4.6, stated live, not
repeated here); an internal check found no evidence that any tested
predictor --- including a hand-built exact reference tracker with full
access to the game's internal state --- reliably beats a uniform prior
at predicting the hidden Regent past the point where its identity is no
longer determined by rule, against this specific opponent; we do not
claim, and this paper does not require, that the underlying task is
demonstrably solvable by any method (see Discussion, §6, on the
tractability question generally).

\textbf{Future work, not a limitation of what is claimed here:} whether
this result is sensitive to elicitation architecture. Joint elicitation,
as used here, may allow the belief request to interact with move
generation; split elicitation avoids persistent-context contamination
but measures belief in a separate model call and context from the one
producing the move. Which better captures the belief associated with the
action is an empirical question --- re-probing a frozen state has itself
been shown to flip an answer 34.8\% of the time in an unrelated
instrument {[}15{]}, so neither architecture should be assumed free of
its own measurement noise. Two further items remain open: whether the
finding holds against a position-responsive (rather than fixed-policy)
opponent, and a systematic multi-seat replication of the full
calibration battery beyond S1.

\begin{center}\rule{0.5\linewidth}{0.5pt}\end{center}

\subsection{6. Discussion}\label{discussion}

\textbf{Why outcome metrics do not surface this.} A model can hold a
badly miscalibrated belief and still win the game the belief was about.
In one representative real game, S1 stated 70\% and 20\% confidence on
two squares and assigned zero probability to a third --- then captured a
piece on that third square for ordinary tactical reasons unrelated to
its stated belief, and that square turned out to hold the opponent's
secretly relocated Regent. The model won. An evaluation that scores only
the outcome --- who won, how many moves, whether the move was legal ---
records this as a clean success. Only an instrument that separately
elicits and scores the stated belief, at a moment distinct from the
action taken on it, can see that the win happened despite the belief,
not because of it. This is not a story about one game or one seat: it is
the same shape as §4.4's within-model dissociation, where disabling
extended reasoning produced a configuration that won on every
conventional metric while producing the worst belief quality tested. One
is a single game a reader can look at directly; the other is a
controlled, within-model comparison across a full sample. Together they
are the paper's central argument stated twice, at two different grains:
an evaluation built only to score outcomes, legality, cost, and latency
will not merely fail to detect poor belief-tracking --- it will actively
reward the configuration that has the least of it.

\textbf{A plausible mechanism, not one we test.} Our result is a
behavioral measurement at the moment of action, not an intervention on
how S1 was trained --- we make no claim about \emph{why} the
miscalibration exists, only that it does and where. It is, however,
consistent with a training-time account proposed elsewhere: reward
models used in RLHF's optimization step have been shown to carry an
inherent bias toward high-confidence verbalizations independent of
actual response quality {[}9{]}, which would push an instruction-tuned
model toward confident wrong beliefs exactly at the moments an evaluator
never separately checks. We note the connection rather than claim it:
nothing here isolates training procedure as the cause, and a base or
non-RLHF'd model on this same instrument is the natural test.

\textbf{Endogenous vs.~exogenous hidden state.} Existing
hidden-information settings used to evaluate language models --- a dealt
hand of cards, an assigned social-deduction role --- give the agent
hidden information it did not choose and cannot change: the task is
purely inference about someone else's fixed fact, or protecting a fact
of one's own by not revealing it. Crown Shift is different in kind: a
player \emph{creates} the hidden fact by choosing where to move it, and
can revise that choice repeatedly over the course of a game, all while
under observation. Concealing an intention you formed and can change is
a different cognitive demand from concealing a card you were dealt once.
This distinction is not incidental to the setting's design --- it is the
reason a coverage gap of the kind measured here (§4.6) is even
expressible: there must be a real decision the agent made about its own
hidden state, not just an inference about someone else's.

\textbf{The oracle-vs-public tractability bound.} A central
methodological risk in any ``the model failed'' claim is that the
underlying task might not have been solvable by any method --- in which
case the finding is really about task design, not model capability. We
tested this directly for our own setting using an oracle (full access to
internal game state) against a public predictor (only what either player
can observe), and found the answer is opponent-policy-dependent: shift
\emph{target}, once a shift is known to have occurred, is substantially
predictable from public position features (83.1\% top-3 accuracy vs.~a
37.5\% naive baseline); shift \emph{timing}, against our fixed opponent,
is close to random by that opponent's own policy, and an oracle with
full internal access gains almost nothing over a public predictor at
forecasting it. A predictor needs both factors jointly, and a near-flat
timing factor collapses the product even when the target factor is well
resolved --- which is the mechanism behind this paper's own finding that
no tested predictor, including a hand-built exact reference tracker,
beats uniform meaningfully past the point where the rules stop
determining the answer, against this opponent. The general point:
distinguishing ``the model failed'' from ``the task was impossible for
anyone'' requires a computable ceiling, and a computable ceiling
requires an opponent whose policy can be read from source. Most
evaluation settings --- anything scored against a closed-source or human
opponent --- cannot make this distinction at all. This is the paper's
most transferable methodological contribution, independent of any single
model's score.

\textbf{Deliberation caps as an evaluation bias.} A fixed token budget
for model output penalizes models that would otherwise deliberate
longer, independent of whether that deliberation would have helped ---
the same ``time vs.~depth'' limitation documented by other game-based
LLM evaluation efforts. We observed this directly: one seat truncated on
roughly a third of its turns at the token budget used throughout this
work, while another almost never did. This cuts in a specific direction
worth naming: it \emph{strengthens} any result showing a capable model
outperforming despite frequent truncation (its deliberation was cut
short and it still did better), and it \emph{weakens} any absolute claim
about a model's calibration ceiling, since a longer budget was never
tested for every seat.

\begin{center}\rule{0.5\linewidth}{0.5pt}\end{center}

\subsection{7. Conclusion}\label{conclusion}

Under an instrument that elicits a stated belief about hidden game state
separately from the action taken on it, and scores both against ground
truth exact and recoverable at reveal, one language model's confidence
at the moment it acts is not merely uninformative: captures made at high
stated confidence were correct in only 1 of 62 cases across two
independent, direction-agreeing batches, worse than an uninformative
prior from the earliest ply at which inference is even possible, present
in weaker form (not claimed universal, and not shown to track external
capability) across four further configurations spanning a second lab.
Irreducible difficulty does not license confidence: where a hidden state
is hard or partly unidentifiable, the appropriate epistemic response is
low or diffuse confidence, not confident error, so
near-unidentifiability makes calibration more important, not less --- we
make no claim that the post-shift state is demonstrably inferable, and
the finding does not require one; no predictor we tested, including a
privileged hand-built reference tracker, reliably beats uniform once the
rules stop determining the answer outright. A model exhibiting this
pattern can still win the game its belief was about, which is why an
evaluation scoring outcomes alone would not surface it. Two extensions
follow directly: whether the same failure holds against an opponent
whose behavior is responsive to the board --- our own fixed opponent's
near-random shift timing is what makes the hidden state unmeasurable
past a certain point here, and a position-responsive opponent is the
natural next instrument --- and whether a score on this platform
predicts anything about model behavior in a deployment setting outside
it, which we consider the more consequential of the two and entirely
unaddressed by this work.

\begin{center}\rule{0.5\linewidth}{0.5pt}\end{center}

\subsection{Use of AI tools}\label{use-of-ai-tools}

This work made extensive use of large language model assistance, and we
describe that use in detail below both to comply with disclosure policy
and because the pattern of failures we encountered may be useful to
others adopting similar workflows.

Three tools were used, in distinct roles, across the project.
\textbf{Claude Opus 5}, used in a conversational chat interface, served
as reviewer and research-direction collaborator: it read drafts and
results, proposed and critiqued framing decisions (including the choice
between a findings-led and an instrument-led paper), identified specific
errors during review passes (including a sentence in §4.5 that stated
the control condition's decline ran in the opposite direction from what
the underlying data showed), ranked what to cut for length, and drafted
the first version of this disclosure section. \textbf{Claude Code}, an
agentic coding assistant operating directly on the project's files and
tools, did the execution: running experiments and analysis code,
verifying every citation against its primary source, writing and editing
all manuscript text and figures under the author's and Opus's direction,
compiling the manuscript to LaTeX to obtain a real page count, and all
git and file operations. Where a fix originated with Opus's review,
Claude Code implemented and re-verified it before it entered the
manuscript; the author approved each round before it proceeded to the
next. \textbf{A third tool, GPT-5.6 Sol, was used later in the process
in a review and literature-search role.} It surfaced three adjacent
works now cited in §2, helped assess their relationship to the paper's
contribution, and reviewed the manuscript's claims and the resulting
correction instructions, including identifying a conceptual error in a
proposed correction before implementation. Its outputs were
independently verified before use: numerical details it supplied for one
cited work were found to be incorrect and were not used. All three tools
are subsequently referred to as ``AI assistance'' below where the
distinction is not load-bearing to the specific claim.

\textbf{The game variant used as this paper's instrument (Regent Chess,
§3) was designed by the author prior to this work.} AI assistance
(Claude, chat interface) was used for its software implementation and to
surface rule edge cases during that implementation. Rule correctness is
established by a test suite covering each rule clause, written with AI
assistance and verified by the suite's execution. We state this
separately from the categories below because it is a distinct kind of
provenance claim: the game's design is prior, independent creative work,
not a research output of this paper or a product of the AI-assisted
process described here --- AI assistance entered only at the
implementation stage, after the design existed.

\textbf{Required-disclosure categories.} We used generative AI tools to:
develop conceptual frameworks (the instrument-design principles in §6
were articulated in dialogue with an AI assistant, though derived from
this work's own empirical findings); propose and refine hypotheses (the
pre-registered expectation branches in §3 and §4.8 were drafted with AI
assistance and ruled on by the author before any corresponding data
existed); design and provide feedback on research methodology and
experiments (the permutation null in §4.6, the population-matched
baseline in §4.1, the paired-bootstrap procedure, the pre-shift
exclusion rule in §3, and the oracle-vs-public tractability analysis in
§6 were all proposed or substantially shaped in dialogue with an AI
assistant); implement methods (all analysis code, the game engine, the
agent harness, and the elicitation pipeline were written with AI coding
assistance under author direction); and interpret results (AI assistance
was used throughout to reason about what each measurement did and did
not support).

We did not use generative AI tools to generate synthetic datasets ---
all game records analysed here are logs of real model play against a
deterministic opponent, at metered API cost. We did not use them to
assist with translation, to formulate survey or interview questions, or
to transcribe research material; these are not applicable to this work.
Mathematical claims in this paper are limited to standard statistical
procedures applied via established libraries; no novel proofs were
formulated or assisted.

\textbf{Recommended-disclosure categories.} We additionally used
generative AI tools to: create and modify all figures; identify and
summarise related literature; format references; suggest the paper's
structure and title; draft all sections of the manuscript against an
author-approved outline; edit for readability; and for general
brainstorming and information search.

\textbf{Verification and responsibility.} All AI-assisted work was
reviewed. Specifically:

\begin{itemize}
\tightlist
\item
  Every headline figure in this paper is produced by version-controlled
  analysis code that writes its output to a file, following an internal
  rule adopted mid-project after a previously-reported result was found
  to have been computed ad hoc and could not be exactly reproduced. A
  single canonical numbers document is the source for every figure cited
  here, and an automated test pins a subset of those figures against the
  underlying committed data.
\item
  Several results reported in earlier internal drafts were withdrawn or
  substantially revised during pre-submission verification, including a
  claimed rank-inversion effect (§4.6), an attribution of
  depth-dependent degradation entirely to hidden-state tracking (§4.5),
  and a claim that the belief-scoring task is demonstrably tractable
  (§5, §6). Each was tested against a pre-registered decision rule and
  reported at the strength the test supported.
\item
  \textbf{Citation verification required multiple passes.} Errors of
  attribution, title, venue, and quotation accuracy were found after
  earlier passes had reported the reference list as verified. The final
  list was checked against the resolved source for each entry,
  confirming that the identifier resolves, that title and authors match,
  and that each source supports the specific claim attributed to it ---
  including a final pass in which the author personally visited primary
  sources himself, unassisted, which located named individual bylines
  the prior AI-driven passes had not surfaced (Oran Kelly and Yao Yan,
  both now cited by name in {[}3{]} and {[}4{]} rather than by
  organization alone). We note this because verifying that a citation
  exists proved insufficient when the reference list was AI-assisted.
\end{itemize}

The author made all substantive research decisions, including the
rulings that resolved ambiguous design questions, the pre-registered
stopping rules and their application, the decision to close data
collection, and the selection of which claims this paper asserts and at
what strength. We take responsibility for the final content of this
work, including all text, claims, and artifacts produced with the aid of
generative AI.

\begin{center}\rule{0.5\linewidth}{0.5pt}\end{center}

\begin{center}\rule{0.5\linewidth}{0.5pt}\end{center}

\subsection{References}\label{references}

{[}1{]} Olszewska, K. and Risdal, M. ``Kaggle Game Arena evaluates AI
models through games.'' \emph{Google Blog --- Innovation and AI}, August
4, 2025.
https://blog.google/innovation-and-ai/products/kaggle-game-arena/

{[}2{]} Kaggle. ``Kaggle Game Arena'' (platform page --- games roster
and methodology). https://www.kaggle.com/game-arena

{[}3{]} Kelly, O. ``Advancing AI benchmarking with Game Arena.''
\emph{Google Blog --- Innovation and AI}, February 2026.
https://blog.google/innovation-and-ai/models-and-research/google-deepmind/kaggle-game-arena-updates/

{[}4{]} Yan, Y. ``Adding Chess Openings to Game Arena.'' \emph{Kaggle
Blog}. https://www.kaggle.com/blog/game-arena-chess-openings

{[}5{]} Kadavath, S., Conerly, T., Askell, A., Henighan, T., Drain, D.,
Perez, E., Schiefer, N., Hatfield-Dodds, Z., DasSarma, N., Tran-Johnson,
E., Johnston, S., El-Showk, S., Jones, A., Elhage, N., Hume, T., Chen,
A., Bai, Y., Bowman, S., Fort, S., Ganguli, D., Hernandez, D., Jacobson,
J., Kernion, J., Kravec, S., Lovitt, L., Ndousse, K., Olsson, C.,
Ringer, S., Amodei, D., Brown, T., Clark, J., Joseph, N., Mann, B.,
McCandlish, S., Olah, C., and Kaplan, J. ``Language Models (Mostly) Know
What They Know.'' arXiv:2207.05221, 2022.
https://arxiv.org/abs/2207.05221

{[}6{]} Kim, H. and Kang, P. ``Same Answer, Different Confidence:
Protocol Sensitivity in LLM Confidence Calibration.'' arXiv:2605.27752,
2026. https://arxiv.org/abs/2605.27752

{[}7{]} Li, Y., Xiong, M., Wu, J., and Hooi, B. ``ConfTuner: Training
Large Language Models to Express Their Confidence Verbally.'' NeurIPS
2025. arXiv:2508.18847. https://arxiv.org/abs/2508.18847

{[}8{]} Wang, V. and Stengel-Eskin, E. ``Calibrating Verbalized
Confidence with Self-Generated Distractors.'' ICLR 2026.
arXiv:2509.25532. https://arxiv.org/abs/2509.25532

{[}9{]} Leng, J., Huang, C., Zhu, B., and Huang, J. ``Taming
Overconfidence in LLMs: Reward Calibration in RLHF.'' arXiv:2410.09724,
2024/2025. https://arxiv.org/abs/2410.09724

{[}10{]} ffrench-Constant, M., Yang, D., Huang, X., and Kapoor, S.
``ConfidenceBench: Evaluating Confidence Calibration in Large Language
Models.'' arXiv:2607.20526, 2026. https://arxiv.org/abs/2607.20526

{[}11{]} Singh, J., Khan, Z., Prasad, A., Chen, J. C.-Y., Nambi, A.,
Lee, H., Stengel-Eskin, E., and Bansal, M. ``Agent-BRACE: Decoupling
Beliefs from Actions in Long-Horizon Tasks via Verbalized State
Uncertainty.'' arXiv:2605.11436, 2026. https://arxiv.org/abs/2605.11436

{[}12{]} Samanta, A., Magesh, A., Lancewicki, T., Jain, A., Yu, Y.,
Sajda, P., Hassani, K., Modi, A., Jiang, D. R., and Efroni, Y.
``BayesBench: Evaluating LLM Belief Trajectories Under Multi-Turn
Evidence Accumulation.'' arXiv:2606.30850, 2026.
https://arxiv.org/abs/2606.30850

{[}13{]} Pal, A., Kitanovski, T., Liang, A., Potti, A., and Goldblum, M.
``Knowing What You Know Is Not Enough: Large Language Model Confidences
Don't Align With Their Actions.'' arXiv:2511.13240, 2025/2026.
https://arxiv.org/abs/2511.13240

{[}14{]} Agarwal, M., Rana, S., Sundoro, T., Berhe, H., Kim, S., Sharma,
V., O'Brien, S., and Zhu, K. ``WOLF: Werewolf-based Observations for LLM
Deception and Falsehoods.'' Spotlight presentation, NeurIPS 2025
Workshop on Multi-Turn Interactions in Large Language Models (MTI-LLM).
arXiv:2512.09187. https://arxiv.org/abs/2512.09187

{[}15{]} Karpov, I. ``MafiaScope: Non-Invasive, Time-Resolved Belief
Probing for LLM Agents in Social Deduction Games.'' arXiv:2607.10645,
2026. https://arxiv.org/abs/2607.10645

{[}16{]} Lin, H.-T. and Hou, T.-Y. ``Readable Minds: Emergent
Theory-of-Mind-Like Behavior in LLM Poker Agents.'' arXiv:2604.04157,
2026. https://arxiv.org/abs/2604.04157

\begin{center}\rule{0.5\linewidth}{0.5pt}\end{center}

\subsection{Appendix: Regent Chess
Rules}\label{appendix-regent-chess-rules}

The rules below reproduce the frozen ``Golden Rules'' ruleset, version
1.2, under which every game reported in this paper was played. Internal
version-history annotations present in the source ruleset have been
removed for readability; no rule text has been altered. Sections of the
source ruleset covering interface requirements, strategic commentary,
and data-handling policy are omitted here, as they do not bear on any
reported result. Rule correctness is established by a test suite in
which each rule clause is covered by at least one test.

\subsubsection{Core principle}\label{core-principle}

Regent Chess is a digital chess variant in which royal status can be
secretly transferred between pieces. The objective is not fixed to the
king: a player wins by capturing the opponent's \emph{Regent}, the piece
that currently holds royal status, which may or may not be that player's
Original King.

\subsubsection{Terminology}\label{terminology}

\begin{itemize}
\tightlist
\item
  \textbf{Regent} --- The piece that currently holds royal status.
  Capturing the opponent's Regent wins the game immediately.
\item
  \textbf{Original King} --- The piece that began the game as the king.
\item
  \textbf{Crown Shift} --- The secret action of transferring royal
  status from the current Regent to another piece.
\item
  \textbf{King Mode} --- A player is in King Mode when their Original
  King is their Regent. Standard chess rules apply to the king.
\item
  \textbf{Regent Mode} --- A player is in Regent Mode when some other
  piece is their Regent. That piece gains royal immunity.
\end{itemize}

\subsubsection{Setup}\label{setup}

The game begins from the standard chess starting position. Both players
start in King Mode with their Original King as Regent. All hidden state
is tracked privately by the software and logged for post-game recovery.

\subsubsection{Crown Shift}\label{crown-shift}

\textbf{Availability.} Crown Shift becomes available to each player at
the start of their 4th turn, that is, after they have completed three
moves. Once used, it becomes available again at the start of the turn
following 15 further completed moves; a player who shifts on their 10th
move may shift again from their 25th. Each player tracks their own move
count independently, and availability is unaffected by the opponent's
actions or timing. Crown Shift may be used even when the Original King
is in check or checkmate.

\textbf{Execution.} Crown Shift may be used only during the player's own
turn. No declaration is made; the action is secret and the opponent is
not notified. The player may select any of their own pieces to become
the new Regent, including the Original King, which returns them to King
Mode. If the Original King has been captured --- possible only while it
was demoted in Regent Mode --- the player can never return to King Mode,
since a captured piece cannot be selected. After executing a Crown
Shift, the player must still make a regular move: the shift does not
consume the turn.

\textbf{Escaping checkmate.} At the start of a turn on which the
Original King is in checkmate, the player may use Crown Shift before
attempting any other move. Assigning royal status to a piece other than
the Original King immediately nullifies the checkmate condition, after
which the player must make a legal move with any piece. If the player
cannot do so, chooses not to shift, or shifts while selecting the
Original King, the checkmate stands and the game is lost.

Timing: if Crown Shift is \emph{unavailable} at the moment checkmate
occurs --- before the 4th turn, or during cooldown --- the game ends
immediately as a loss. If it is \emph{available}, the game pauses for
that player's turn so the escape may be attempted.

\textbf{Risk.} Royal status may be shifted to any piece, including one
currently under attack, in which case it may be captured immediately and
the game lost. This includes shifting away from an Original King that is
itself under attack.

\subsubsection{Game modes and piece
behaviour}\label{game-modes-and-piece-behaviour}

\textbf{King Mode.} Standard chess rules apply to the Original King. It
cannot move into or remain in check; if it is in check, the player must
address the threat by moving, blocking, or capturing. Castling is
permitted subject to standard restrictions, and is possible only in King
Mode.

\textbf{Regent Mode --- the Regent.} The Regent moves according to the
standard rules for its piece type, is immune to all check restrictions,
and may move into or remain on attacked squares. Its capture ends the
game immediately.

\textbf{Regent Mode --- the demoted Original King.} The Original King
retains its movement pattern of one square in any direction but is no
longer royal. It may move into or remain on attacked squares, and may be
deliberately exposed as a sacrifice. It cannot castle. Its capture is
material loss only and does not end the game.

\subsubsection{Win conditions}\label{win-conditions}

A player wins immediately upon capturing the opponent's Regent; the game
ends at the instant the capture is executed.

Checkmate applies only in King Mode: if the Original King as Regent is
checkmated and the player cannot or does not use Crown Shift to escape,
the game is lost. In Regent Mode traditional checkmate is impossible,
since the Regent is immune to check restrictions.

\subsubsection{Special rules and
interactions}\label{special-rules-and-interactions}

\textbf{Pawn promotion.} If a Regent pawn promotes, the promoted piece
automatically becomes the new Regent, gaining both the movement of its
new type and royal immunity. The transition is silent.

\textbf{Check announcements.} The software announces check whenever a
player's Original King is attacked, regardless of whether that player is
in King Mode or Regent Mode. In King Mode the player must respond; in
Regent Mode the player may ignore it and move freely, though responding
anyway may preserve deception. Because the announcement does not depend
on the player's mode, it functions as a probing signal for the opponent.

\textbf{Stalemate and passing.} Traditional stalemate is abolished. A
player with no legal move available has their turn automatically passed;
play continues with the opponent and no draw results from this alone.

In King Mode this applies only when the player is not in check --- a
King-Mode player in check with no legal move is checkmated, or
checkmate-pending, instead. In Regent Mode it applies regardless of
whether the Original King is under attack, since Regent Mode already
permits remaining in check.

A pass does not count as one of the player's own moves for Crown Shift
purposes: it advances neither the eligibility threshold nor the
cooldown. It does not count toward the 50-move rule below, which
measures player decisions, but it does count toward the 75-move
backstop, which measures elapsed plies. A pass forfeits nothing beyond
the turn: a player who had Crown Shift available before passing still
has it on their next real turn. Crown Shift cannot be attempted on a
passed turn, since passing means no action was available at all.

A pass is not announced as a separate event but is visible in the game
record as a gap in the normal alternation of moves. If both players pass
on consecutive turns, the game is an immediate draw.

\textbf{Forced draw conditions.} A game is drawn if any of the following
occur.

\begin{enumerate}
\def\labelenumi{\arabic{enumi}.}
\tightlist
\item
  \textbf{Consecutive passes.} Both players pass consecutively, having
  no legal moves.
\item
  \textbf{50-move rule.} Fifty consecutive half-moves have been made
  without a capture and without a Crown Shift. Captures and Crown Shifts
  each reset this counter. Passed turns do not count toward it.
\item
  \textbf{75-move backstop.} Seventy-five full moves without a capture
  or pawn move. This counter is \emph{not} reset by Crown Shifts, and
  passed turns do count toward it. The asymmetry with the 50-move rule
  is deliberate: the 50-move rule forces meaningful decisions, while the
  backstop bounds game length regardless of why a ply occurred,
  preventing a losing player from stalling indefinitely by shifting
  every 15 moves.
\item
  \textbf{Threefold repetition.} The same \emph{visible} board state ---
  piece placement, side to move, castling rights, en passant
  availability --- occurs three times. Hidden state, that is the
  identity of either Regent, is not part of the comparison; a secret
  Crown Shift between occurrences does not prevent the draw.
\item
  \textbf{Dead position.} Only the two Original Kings remain.
\end{enumerate}

Insufficient-material draws are otherwise abolished. Standard chess
declares positions such as king and knight versus king drawn because
checkmate is impossible, but in Regent Chess a lone knight can still win
by capturing the Regent. Any material beyond bare kings keeps the game
live.

\textbf{Castling restrictions.} Castling is permitted only in King Mode
and must involve the Original King as Regent together with a rook. A
rook that is the current Regent cannot be used to castle. All standard
castling restrictions apply. Returning to King Mode does not restore
castling rights lost while demoted: if the Original King or the relevant
rook moved at any point, castling with it remains permanently
unavailable.

\subsubsection{Standard chess rules}\label{standard-chess-rules}

All other standard chess rules remain in effect, including en passant,
piece movement patterns except for the check immunities described above,
and turn alternation.

\end{document}